\documentclass[journal, twoside]{IEEEtran}
\IEEEoverridecommandlockouts
\usepackage{lipsum}
\usepackage{hyperref}
\usepackage{placeins}
\usepackage{mathtools}
\usepackage{graphicx}
\usepackage{float}
\usepackage{setspace}
\usepackage{color}
\usepackage[dvipsnames]{xcolor}
\usepackage{cite}
\usepackage{indentfirst}
\usepackage{amssymb}
\usepackage{amsmath}
\usepackage{amsbsy}
\usepackage{subfigure}
\usepackage{graphicx}
\usepackage{setspace}
\usepackage{soul}
\usepackage{algorithm}
\usepackage{algorithmic}
\usepackage[ruled,linesnumbered,algo2e]{algorithm2e}
\usepackage{bm}

\begin{document}

\title{Physics-informed Neural Network Predictive Control for Quadruped Locomotion}

\author{Haolin Li, Yikang Chai, Bailin Lv, Lecheng Ruan, Hang Zhao, Ye Zhao, Jianwen Luo\textsuperscript{$*$}
\vspace{-5mm}
\thanks{This work was supported in part by National Natural Science Foundation of China under Grant 51905251, the Shenzhen Major Science and Technology Program under Grant 202402004. \textit{(Corresponding author: Jianwen Luo. Email: luojw76@mail.sysu.edu.cn)}}
\thanks{Haolin Li, Bailin Lv and Jianwen Luo are with the School of Intelligent Systems Engineering, Shenzhen Campus of Sun Yat-sen University, Shenzhen, Guandong 518063, China, and also with General Embodied AI Center of Sun Yat-sen University, and also with Guangdong Provincial Key Laboratory of Fire Science and Intelligent Emergency Technology, Guangzhou 510006, China. (email: haolin.li.2000@gmail.com).}
\thanks{Yikang Chai is with General Embodied AI Center of Sun Yat-sen University. (email: kang.yi622@gmail.com)}
\thanks{Lecheng Ruan is with College of Engineering, Peking University, Beijing, 100871, China. (email: ruanlecheng@ucla.edu) }
\thanks{Hang Zhao is with Robotics and Autonomous Systems Thrust, The Hong Kong University of Science and Technology (Guangzhou), Guangzhou, China. (email: hangzhao@hkust-gz.edu.cn)}
\thanks{Ye Zhao is with the George W. Woodruff School of Mechanical Engineering, Georgia Institute of Technology, Atlanta, USA. (email: ye.zhao@me.gatech.edu)}
}

\markboth{Manuscript has been submitted for review}
{Li  \MakeLowercase{\textit{et al.}}: Physics-informed Neural Network Predictive Control for Quadruped Locomotion}

\maketitle

\begin{abstract}
This study introduces a unified control framework that addresses the challenge of precise quadruped locomotion with unknown payloads, named as online payload identification-based physics-informed neural network predictive control (OPI-PINNPC). By integrating online payload identification with physics-informed neural networks (PINNs), our approach embeds identified mass parameters directly into the neural network's loss function, ensuring physical consistency while adapting to changing load conditions. The physics-constrained neural representation serves as an efficient surrogate model within our nonlinear model predictive controller, enabling real-time optimization despite the complex dynamics of legged locomotion. Experimental validation on our quadruped robot platform demonstrates 35\% improvement in position and orientation tracking accuracy across diverse payload conditions (25-100 kg), with substantially faster convergence compared to previous adaptive control methods. Our framework provides a adaptive solution for maintaining locomotion performance under variable payload conditions without sacrificing computational efficiency.
\end{abstract}
\vspace{-2 mm}
\section{Introduction}

\IEEEPARstart{N}{onlinear} Model Predictive Control (NMPC) has emerged as a powerful and versatile control strategy, particularly in the realm of legged locomotion \cite{10286076, 10669220, 10138309, 9636262}. Quadruped robots, with their complex dynamics and multiple degrees of freedom, present unique challenges for motion control. Traditional control methods often struggle to meet these demands due to the highly nonlinear and coupled nature of quadruped dynamics \cite{9561913, 9321699, 9812433}. NMPC addresses these challenges by leveraging a predictive model of the system to optimize control inputs over a finite horizon \cite{9560976, 9113252}. Unlike linear control approaches \cite{luo2019}, NMPC fully embraces the nonlinearities inherent in quadruped dynamics, leading to more accurate and effective control strategies \cite{9361258}. It considers the future states and constraints of the robot, enabling it to anticipate and react to changes in terrain, disturbances, or desired trajectories in real-time \cite{10008229}. This predictive capability is crucial for tasks such as dynamic walking, running, and jumping, where the robot must adapt rapidly to maintain stability and achieve its objectives \cite{10801676}. Compared with analytical methods \cite{luo-ral, luo-tase}, NMPC incorporates constraints like joint limits, friction, and contact forces into optimization, ensuring control inputs are both optimal and safe for robot's hardware \cite{10160507}. NMPC's flexibility enables adaptation to different quadruped designs and operational scenarios \cite{ijrr_luo, frobt724138}. Recent advances in computing and optimization have made NMPC practical for real-time controls. However, accurately modeling nonlinear dynamics in NMPC constraints remains challenging under uncertainties or disturbances, especially with the unknown parameters of the locomotion model under varying payloads \cite{10705076, jin2022unknown}.

As robotic systems have grown increasingly complex, both paradigms face distinct limitations: traditional machine learning models typically sacrifice physical interpretability and consistency, while purely physics-based approaches struggle to adapt to real-time variations and uncertainties \cite{ramp-net, sanyal2024ev, yang2023collaborative}. Physics-informed Neural Networks (PINNs) have emerged as a groundbreaking methodology in robot modeling and control, effectively bridging the gap between conventional physics-based techniques and contemporary machine learning \cite{raissi2019physics,nicodemus2022physics,bensch2024physics}.


PINNs overcome these limitations by embedding fundamental physical laws into the architecture and training process of neural networks. This integration enables PINNs to leverage prior knowledge of physics, reducing the dependency on large datasets and improving model generalizability\cite{ramp-net, luo2024research}. By embedding robotic system dynamics into their structure, PINNs can accurately predict responses and identify dynamic parameters, even with nonlinearities and dynamic coupling. Moreover, PINNs show great potential in control applications. Recent research has introduced frameworks such as Physics-Informed Neural Nets for Control (PINC)\cite{ANTONELO2024127419}, which extend PINNs by integrating with model predictive control and other control techniques. The framework enables real-time simulation and control of complex robotic systems, outperforming traditional numerical methods in computational efficiency and prediction accuracy.

This study proposes a method to approximate the nonlinear dynamics of quadruped locomotion via physics-informed neural network with an unknown payload identification algorithm. Three contributions are highlighted below:

\begin{enumerate}
\item We propose online payload identification-based physics-informed neural network predictive control (OPI-PINNPC), a novel control architecture that integrates online payload identification, physics-informed neural network (PINN), and nonlinear model predictive control (NMPC), establishing a unified framework for addressing quadruped locomotion tracking control in the presence of unknown payloads.
\item We incorporate the identified payload mass and orientation parameters as critical inputs to the physics-informed loss function of PINN, enabling accurate prediction of quadruped robot dynamics with unknown payload through physical principle constraints.
\item Compared with our previous work on adaptive control algorithm for quadruped locomotion (ACQL) \cite{jin2022unknown}, the proposed OPI-PINNPC demonstrates faster convergence rate and higher tracking accuracy through experiments, while maintaining superior adaptation across payload variations (25-100 kg).
\end{enumerate}

\section{Problem Formulation}
This section introduces the problem formulation, including the nonlinear dynamics of quadruped locomotion, corresponding nonlinear model predictive control and physics-informed neural network.
\vspace{-0.2cm}
\subsection{Quadruped Locomotion Dynamics}
\begin{figure*}[ht]
    \centering
    \setlength{\abovecaptionskip}{ 0 cm}
    \setlength{\belowcaptionskip}{-5 cm}
    \includegraphics[width=0.7\linewidth]{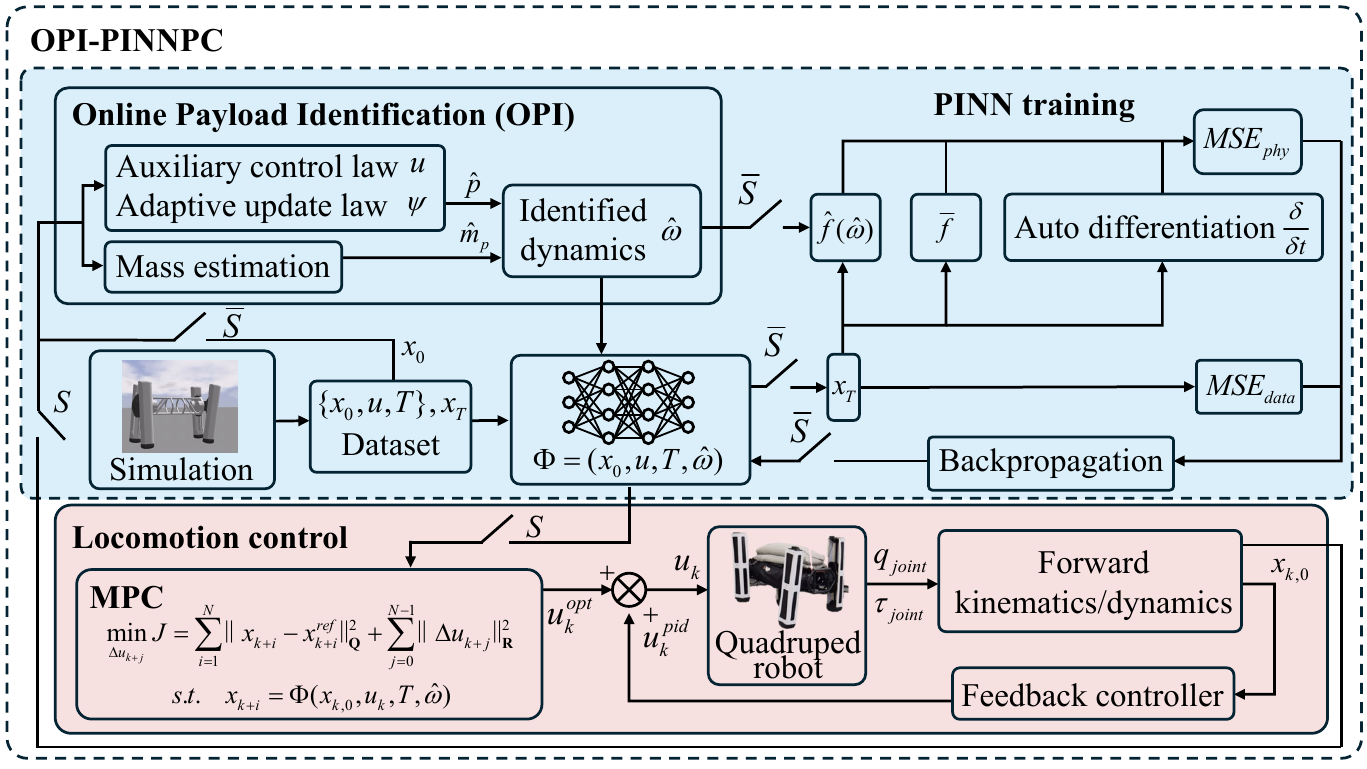}
    \caption{The overall framework of OPI-PINNPC. (i) Training phase (switch $\bar S$ closed): OPI module estimates the unknown payload parameters via the input data. The identified parameters subsequently construct the physics-informed loss, which is combined with the data loss computed by labeled data to optimize the network weights through backpropagation. (ii) Prediction phase (switch $S$ closed): OPI synthesizes payload parameters using real-time robot states, followed by the PINNs state prediction fulfilling NMPC integration requirements, ultimately generating composite control inputs consisting of the optimal control via NMPC optimization and the feedback control.}
    \label{fig:block_diagram}
\end{figure*}

We use Kirin, a specifically designed quadruped robot featuring electrically actuated prismatic knee joints \cite{luo2022prismatic}, as the platform for the demonstration of our proposed method, as shown in Fig. \ref{fig:block_diagram}. The nonlinear dynamics is described as follows:
\begin{equation}
\label{eq:quadruped_locomotion}
    \begin{cases}
        m\ddot{r}&=\sum F_{ci} - \sum m_i g - m_p g \\
        I\ddot{\theta}&=\sum r_{ci} \times F_{ci} + \sum r_{i} \times m_i g + r_p \times m_p g
    \end{cases},
\end{equation}

\noindent where $m$ and $r \in R^3$ denote the total mass of the quadruped robot and the position of quadruped locomotion, respectively. $F_{ci} \in R^3$ denotes the contact force of foot $i$. $m_i$ and $m_p$ denote the mass of leg $i$ and the payload to be identified, respectively. $g$ denotes gravitational acceleration, $I \in R^{3\times3}$ denotes rotational inertia matrix of the robot. $\theta \in R^3$ denotes the angular position of the robot torso. $r_{ci} \in R^3$, $r_i \in R^3$ and $r_p \in R^3$ denote the position vector from the forces $F_{ci} \in R^3$, $m_i g$ and $m_p g$ with respect to the Center of Mass, respectively. The symbol notations are the same as \cite{luo2022prismatic}.
\vspace{-0.3cm}
\subsection{Nonlinear Model Predictive Control}

Nonlinear dynamics in Eq.~\eqref{eq:quadruped_locomotion} is typically expressed in the general control system form as below:
\begin{equation}
\label{eq:continuos_system}
    \dot{x} = f(x,u),
\end{equation}

\noindent where $x=x(t)$ denotes the state variables, $u=u(t)$ denotes the control input. The dynamics function $f(x,u)$ is assumed to be Lipschitz-continuous. Eq.~\eqref{eq:quadruped_locomotion} is usually formulated into the discrete form as rewritten below:
\begin{equation}
\label{eq:discrete_system}
    x_{k+1} = g(T,x_{k},u_{k}),
\end{equation}

\noindent where $x_{k}$ and $u_{k}$ denote the state variables and control input at time instant $k$, and $T$ is the sampling time interval.

Given $x^{\rm ref}_{k+1},i=1,2,...,N$ as the reference trajectory sequence of system \eqref{eq:discrete_system} at time instant $k$ with a prediction horizon $N$. The NMPC can be formulated as:

\begin{equation}
\label{eq:qp_formulation}
\min_{\Delta u_{k+j}} 
\sum_{i=1}^{N} \Vert x_{k+i} - x^{\rm ref}_{k+i} \Vert_{{\textit{\textbf{Q}}}}^2 + \sum_{j=0}^{N-1} \Vert \Delta{u_{k+j}} \Vert_{\textit{\textbf{R}}}^2
\end{equation}

\begin{equation}
\label{eq:mpc_constrains}
\begin{aligned}
&\text{s.t.} \quad 
\begin{cases}
x_{k+i+1} = g(T,x_{k+i},u_{k+i}) \\
u_{k+i+1} = u_{k+i} + \sum_{j=0}^{i} \Delta u_{k+j} \\
u_{\min} \leq u_{k+i} \leq u_{\max} \\
\Delta u_{\min} \leq \Delta u_{k+i} \leq \Delta u_{\max} \\
x_{\min} \leq x_{k+i} \leq x_{\max}\\
\end{cases} \\
&\quad \quad \quad \quad \quad \quad \forall i = 0,\ldots,N-1,
\end{aligned}
\end{equation}

\noindent where $\Vert * \Vert_{{\textit{\textbf{M}}}} = \sqrt{*^T {{\textit{\textbf{M}}}} *}$. And $\Delta u_k$ represents the control input increment at at time instant $k$, $\textit{\textbf{Q}}$ and $\textit{\textbf{R}}$ are diagonal matrices that weight the penalization of tracking error and control input increments. 

A primary challenge in solving the above NMPC problem is the computational complexity and numerical precision inherent in numerical integration processes, specifically the iterative computation described by $x_{k+i+1} = g(T,x_{k+i},u_{k+i})$ in Eq.~\eqref{eq:mpc_constrains}. To mitigate this computational burden, this study employs the PINNs framework to capture the underlying nonlinear dynamics and enable rapid state prediction through neural network approximation.

\vspace{-0.5 cm}
\subsection{Physics-Informed Neural Networks}

PINNs is a novel deep learning framework that combines data-driven approaches with fundamental physical principles. It excels at modeling dynamic systems governed by ordinary differential equation (ODE) and predicting system states. Consider the ODE in Eq. \eqref{eq:discrete_system}, the primary objective of PINNs lies in constructing a function \bm{$\Phi$} (typically a neural network) that approximates the dynamics of Eq. \eqref{eq:discrete_system}, that is,
\begin{equation}
x_{k+1} \approx \hat x_{k+1} = \bm{\Phi}(T,x_{k},u_{k}).
\end{equation}

The loss function of PINNs typically consists of two components, which are given by mean squared error (MSE):
\begin{equation}
        \label{eq:loss_func}
        MSE = MSE_{\rm data} + MSE_{\rm phy},
\end{equation}

\noindent where the first component $MSE_{\rm data}$ represents the data loss function that evaluates the deviation between the output from PINNs and labeled data as the supervised part. $MSE_{\rm data}$ is defined as follows:
\begin{equation}
    MSE_{\rm data} = \frac{1}{N_{\rm data}} \sum_{i=1}^{N_{\rm data}} \lVert \hat x_i(v_i) - x_{i} \rVert^{2},
\end{equation}

\noindent where $v_i$ and $x_i$ denote the input and the true state of the dynamic system of the $i$th labeled data from dataset $\{(v_i, x_i): i=1,2,...,N_{\rm data}\}$, respectively. $N_{\rm data}$ is the number of labeled data samples. $\hat x_i(v_i)$ denotes the output from PINNs with $v_i$ as the input. For Eq.~\eqref{eq:discrete_system}, we define the symbol $\bm{\mathcal{G}}(x)$ as:
\begin{equation}
    \bm{\mathcal{G}}(x_k) := \bm{\dot \Phi}(T,x_{k},u_{k}) - g(\tau,x_k,u_k).
\end{equation}

The second component of the loss function Eq. \eqref{eq:loss_func} that distinguishes PINNs from other conventional neural networks is the physics-informed loss function $MSE_{\rm phy}$:
\begin{equation}
    MSE_{\rm phy} = \frac{1}{N_{\rm phy}} \sum_{i=1}^{N_{\rm phy}} \lVert \bm{\mathcal{G}}(\hat x_{i}) \rVert^{2},
\end{equation}

\noindent where $N_{\rm phy}$ denotes the number of unlabeled data. Similar to \cite{ramp-net, antonelo2024physics}, PINNs operate through random sampling of the unlabeled collocation points, where the physics-informed loss function eliminates the requirement for labeled ground truth data, endowing PINNs with a partially unsupervised learning paradigm.
\vspace{-0.4cm}

\section{Methods}
During practical implementation of quadruped locomotion, two primary categories of uncertainties typically emerge: parametric uncertainties arising from component aging or unknown payload variations and non-parametric uncertainties stemming from external disturbances.

This study primarily investigates the quadruped locomotion problem under unknown payload conditions. It is well-known that the NMPC formulation requires an accurate dynamic model for numerical integration, and the computational complexity increases with the nonlinearity of the system. In fact, obtaining an accurate dynamic model for quadruped locomotion is challenging in the presence of an unknown payload. To address the challenges, this study proposes Online Payload Identification-based Physics-Informed Neural Network Predictive Control (OPI-PINNPC) for quadruped locomotion with payload variation.
\vspace{-0.4cm}
\subsection{Architecture of OPI-PINNPC}

As illustrated in Fig.~\ref{fig:block_diagram}, OPI-PINNPC incorporates four components: (i) an online payload identification (OPI) module that estimates payload properties, (ii) a PINNs architecture that generates state predictions through learned system dynamics, (iii) a modified NMPC procedure that replacing numerical integration implementation with PINNs inference, and (iv) a composite controller combining the feedback and feedforward control that deployed for quadruped locomotion. The proposed method aims to address quadruped locomotion under unknown payload conditions through real-time payload identification while establishing an efficient predictive control framework via PINNs.

Notably, the explicit incorporation of payload parameters as inputs distinguishes this work from PINNs-based control frameworks in the existing literature. The objective of this work is to establish a mapping between payload variations and the quadruped robot’s dynamics through the PINNs architecture. To achieve this, the OPI module is embedded into the PINNs framework, where the identified payload parameters formulate the physics-informed loss, enabling the neural network to learn the intrinsic relationship between payload parameters and robot dynamics. This design ensures that the trained PINNs capture payload-dependent nonlinearity, thereby enabling accurate prediction of the robot’s state variables under arbitrary payload conditions. 
\vspace{-0.5cm}
\subsection{Formulation and Algorithm of OPI-PINNPC }

Consider an unknown payload, Eq.~\eqref{eq:quadruped_locomotion} is reformulated as follows:
\begin{equation}
\label{eq:second_order_state_space}
\begin{aligned}
{\begin{bmatrix}
{{{\ddot r}_b}}\\
{{{\ddot \theta }_b}}
\end{bmatrix}} =& {\begin{bmatrix}
{\frac{1}{m}{{\textit{\textbf{I}}}_3}}&{{{\textit{\textbf{O}}}_{3 \times 3}}}\\
{{{\textit{\textbf{O}}}_{3 \times 3}}}&{{{\textit{\textbf{I}}}^{ - 1}}}
\end{bmatrix}} {\begin{bmatrix}
{\sum {{F_{ci}}}  - \sum {{m_i}} g}\\
{\sum {{r_{ci}}}  \times {F_{ci}} + \sum {{r_i}}  \times {m_i}g}
\end{bmatrix}}  \\ 
&+ {\begin{bmatrix}
{\frac{1}{m}{{\textit{\textbf{I}}}_3}}&{{{\textit{\textbf{O}}}_{3 \times 3}}}\\
{{{\textit{\textbf{O}}}_{3 \times 3}}}&{{{\textit{\textbf{I}}}^{ - 1}}}
\end{bmatrix}} {\begin{bmatrix}
{ - {m_p}g}\\
{{r_p} \times {m_p}g}
\end{bmatrix}} .
\end{aligned}
\end{equation}
where we define $x_1 = [x_{11}^T, x_{12}^T]^T = [r^T, \theta^T]^T$, $x_2 = [x_{21}^T, x_{22}^T]^T = [\dot r^T, \dot \theta^T]^T$ and $x = [x_1^T, x_2^T]^T$, Eq. \eqref{eq:second_order_state_space} can be rewritten in the following standard state-space form:

\begin{equation}
\label{eq:state_space}
\dot x 
= \begin{bmatrix}{{{\textit{\textbf{O}}}_{6 \times 6}}}&{{{\textit{\textbf{I}}}_6}}\\{{{\textit{\textbf{O}}}_{6 \times 6}}}&{{{\textit{\textbf{O}}}_{6 \times 6}}}\end{bmatrix}x 
+ \begin{bmatrix}{{{\textit{\textbf{O}}}_6}}\\{{{\textit{\textbf{A}}}_1}{{\bar f}_1}}\end{bmatrix} 
+ \begin{bmatrix}{{{\textit{\textbf{O}}}_6}}\\{{{\textit{\textbf{A}}}_1}{{\hat f}_1}}\end{bmatrix} ,
\end{equation}

\noindent where ${\textit{\textbf{A}}}_1$, $\bar f_1$ and $ \hat f_1$ are given by:

\begin{equation}
\label{eq:state_space_abbreviations}
\begin{aligned}
\begin{cases}
{\textit{\textbf{A}}}_1 = \begin{bmatrix}{\frac{1}{m}{{\textit{\textbf{I}}}_3}}&{{{\textit{\textbf{O}}}_{3 \times 3}}}\\{{{\textit{\textbf{O}}}_{3 \times 3}}}&{{{\textit{\textbf{I}}}^{ - 1}}}\end{bmatrix}\\ \\
 \bar f_1 = \begin{bmatrix}{\sum {{F_{ci}}}  - \sum {{m_i}} g}\\{\sum {{r_{ci}}}  \times {F_{ci}} + \sum {{r_i}}  \times {m_i}g}\end{bmatrix}\\ \\
 \hat f_1 = \begin{bmatrix}{ - {m_p}g}\\{{r_p} \times {m_p}g}\end{bmatrix}
 \end{cases}.
\end{aligned}
\end{equation}

Due to the presence of unknown payload, the latter term $\hat f_1$ of Eq. \eqref{eq:state_space_abbreviations} is to be identified. Building upon our previous investigations in adaptive control for quadruped locomotion \cite{jin2022unknown}, where a real-time quadruped locomotion control framework based on online payload identification (denoted as ACQL) is established, this study extracts the online payload identification algorithm of ACQL to identify the unknown terms $m_p$ and $r_p \times m_p g$ in Eq. \eqref{eq:quadruped_locomotion}.

The identification of $\hat{m}_p$ is the same as \cite{jin2022unknown}, which is relatively straightforward. To estimate ${{r_p} \times {m_p}g}$, the second row formulation of Eq.\eqref{eq:second_order_state_space} is rewritten as follows:

\begin{equation}
\label{eq:identify_d}
\left\{
\begin{array}{l}
\dot{x}_{12} = x_{22} \\
\dot{x}_{22} = u + k + p\\
\dot{\hat p} = \psi
\end{array}
\right.,
\end{equation}

\noindent where $u = I^{-1} \sum r_{ci} \times F_{ci}$ denotes the auxiliary control input, $k = I^{-1} \sum r_{i} \times m_i g$, $p = I^{-1} r_{p} \times m_p g$ denotes the unknown parameter to be identified. $\hat p$ denotes the estimate value of $p$, $\psi$ denotes the adaptive update law.

For the given reference tracking signal of torso orientation $x_{12}^{\rm ref}$, which is a constant during the identification process, as the torso is constrained to vertical-axis motion while maintaining invariant orientation, let $e_1 = x_{12} - x_{12}^{\rm ref}$, we have $\dot e_1 = \dot x_{12} = x_{22}$. Introduce the auxiliary error variable $\bar e = \dot e_1 + {\textit{\textbf{W}}} e_1$ where ${\textit{\textbf{W}}}$ is a positive definite matrix, we have:
\begin{equation}
    \dot{\bar e} = u + k + p + {\textit{\textbf{W}}} x_{22}.
\end{equation}

Then, the control law and adaptive update law is proposed:
\begin{equation}
\label{eq:auxiliary_control_law}
u = - k - \hat p - {\textit{\textbf{W}}} x_{22} + {\textit{\textbf{V}}} r ,
\end{equation}
\vspace{-0.1cm}
\begin{equation}
\label{eq:update_law}
\left\{
\begin{array}{l}
\psi = - {\textit{\textbf{K}}} (u + \hat{p} + \Psi(x_{12}, x_{22}) + k) \\
\dot{z} = - {\textit{\textbf{K}}} z
\end{array}
\right.,
\end{equation}

\noindent where ${\textit{\textbf{V}}}$ is a positive definite matrix, ${\textit{\textbf{K}}}$ denotes the gain diagonal matrix, $\Psi(x_{12}, x_{22})$ is chosen as ${\textit{\textbf{K}}} x_{22}$, $z = \hat p - p + \Psi(x_{12}, x_{22})$.

It has been proven that with the action of auxiliary control law \eqref{eq:auxiliary_control_law} and update law\eqref{eq:update_law}, the estimated value $\hat p$ is able to converge to the true value $p$, and the auxiliary error variable $\bar e$ converges to zero as well \cite{jin2022unknown}.

With the identified $\hat m_p$ and $\hat p$, define $\hat \omega = [\hat m_p, \hat p^T]^T$ and the identified dynamics of system~\eqref{eq:state_space} can be described as follows:
\vspace{4px}
\begin{equation}
\label{eq:abbreviated_state_space}
\dot x  = \bar f(x,u) + \hat f(\hat \omega),
\end{equation}

\noindent where $\bar f(x,u)$ is given by:

\begin{equation}
\bar f(x,u)= \begin{bmatrix}{{{\textit{\textbf{O}}}_{6 \times 6}}}&{{{\textit{\textbf{I}}}_6}}\\{{{\textit{\textbf{O}}}_{6 \times 6}}}&{{{\textit{\textbf{O}}}_{6 \times 6}}}\end{bmatrix}x + \begin{bmatrix}{{{\textit{\textbf{O}}}_6}}\\{{{\textit{\textbf{A}}}_1}{{\bar f}_1}}\end{bmatrix}.
\end{equation}
And $\hat f(\omega)$ is given by:

\begin{equation}
    \hat f(\hat \omega) = \begin{bmatrix}{{{\textit{\textbf{O}}}_6}}\\{{{\textit{\textbf{A}}}_1}{{\hat f}_1}}\end{bmatrix}.
\end{equation}
 
$\bar f(x,u)$ and $\hat f(\hat \omega)$ denote the nominal dynamics and the identified dynamics of quadruped locomotion, respectively.

Discretize the system \eqref{eq:abbreviated_state_space}, resulting in the discrete-time model:

\begin{equation}
\label{eq:discrete_state_space}
    x_{k+1} = \bar g(T,x_k,u_k) + \hat g(\hat\omega).
\end{equation}

In summary, the true dynamics governing quadruped locomotion consists of two components: (i) the nominal dynamics $\bar g(T,x_k, u_k)$ and (ii) the unknown payload dynamics $\hat g(\omega)$ w.r.t. the unknown payload parameters $\omega$. In this study, the PINNs methodology is employed to construct an accurate approximation of the true system dynamics, formulated as: 

\begin{equation}
\label{eq:approx_dynamics}
\Phi(x_k, u_k,T,\hat \omega) \approx \bar g(T,x_k, u_k) + \hat g(\omega),
\end{equation}

\noindent where $\hat \omega$ is the identified payload information. The data loss function is defined as:
\begin{equation}
\begin{aligned}
    MSE_{\rm data} = \frac{1}{N_{\rm data}} \sum_{i=1}^{N_{\rm data}} \lVert \Phi(x_k, u_k,T,\hat \omega) - x_{k+1} \rVert^{2}.
\end{aligned}
\end{equation}

The labeled dataset is gathered from simulation with input-output pairs $(\{x_k, u_k, T \} ,x_{k+1})$ (see Section \ref{sec:PINN_Training}). The inputs are associated with measurable state variables $x_k$, control inputs $u_k$ and sampling time $T$. The output is the corresponding state variables of the next instant $x_{k+1}$. Note that the identified payload parameters $\hat \omega$ is also integrated as inputs to enhance the generalization of neural networks. 

The identified payload information $\hat \omega$ is subsequently incorporated into the physics-informed loss function to derive the true dynamics of quadruped locomotion, thereby enforcing physical consistency in the PINNs predictions. The physics loss function is defined as follows:

\begin{equation}
\begin{aligned}
    MSE_{\rm phy} = \frac{1}{N_{\rm phy}} \sum_{i=1}^{N_{\rm phy}} \lVert \dot \Phi(x_k, u_k,T,\hat \omega)
    & \\ - (\bar g(T,x_k, u_k) + \hat g(\hat \omega)) \rVert^{2}.
\end{aligned}
\end{equation}

Based on data and physics loss functions $MSE_{\rm data}$ and $MSE_{\rm phy}$, the PINNs is trained to capture the true dynamics of quadruped locomotion, and this dual-losses learning method allows the network to establish an accurate state prediction as follows: 
\vspace{-0.1cm}
\begin{equation}
x_{k+1} = \Phi(x_k, u_k,T,\hat \omega) \approx \bar g(T,x_k, u_k) + \hat g(\omega).
\end{equation}

The learned predictive model subsequently replaces conventional numerical integration in the NMPC framework, which is reformulated as follows:

\begin{algorithm}[!t]
\caption{OPI-PINNPC Algorithm}   
\label{alg_balance_control}  
\renewcommand{\algorithmicrequire}{\textbf{Input:}}
\renewcommand{\algorithmicensure}{\textbf{Output:}}
\begin{algorithmic}[1]
{\fontsize{8pt}{12pt}\selectfont
\REQUIRE $N$, $T$, $q_{\rm joint}$, $\dot q_{\rm joint}$, $\tau_{\rm joint}$, $m_i$, $\{x_k^{\rm ref},k=1,2,...,N_{t}\}$
\ENSURE Composite control $u_k$
\STATE Hyperparameter initialization: $k=1$, $e_{\rm threshold}$, ${\textit{\textbf{Q}}}$, ${\textit{\textbf{R}}}$, ${\textit{\textbf{K}}}_p$, ${\textit{\textbf{K}}}_i$, ${\textit{\textbf{K}}}_d$
\STATE $\theta, r \Leftarrow$ Forward kinematics$(q_{\rm joint}, \dot q_{\rm joint})$
\STATE $F \Leftarrow$ Forward dynamics$(\tau_{\rm joint})$
\STATE $\hat m_p \Leftarrow \ddot r, m_i, F$ 
\WHILE {$e > e_{\rm threshold}$}
\STATE $u \Leftarrow$ Auxiliary control law \eqref{eq:auxiliary_control_law}
\STATE $\psi \Leftarrow$ Update law \eqref{eq:update_law}
\STATE $\hat p \Leftarrow u, \psi$
\ENDWHILE
\STATE $\hat \omega \Leftarrow \hat m_p, \hat p$
\WHILE{$k < N_t$}
\STATE $x_{k+1} \Leftarrow $ Forward kinematics$(q_{\rm joint}, \dot q_{\rm joint})$
\FOR{$i = 0,1,...,N$}
\STATE $x_{k+i+1} \Leftarrow \Phi(x_{k+i}, u_{k+i}^{\rm opt},T,\hat \omega)$ 
\STATE $u_{k+i}^{\rm opt} \Leftarrow$ Minimize \eqref{eq:modified_cost_function} s.t. \eqref{eq:modified_constrains} via OSQP
\ENDFOR
\STATE $u_k \Leftarrow$ Composite controller
\STATE $k \Leftarrow k+1$
\ENDWHILE
}
\end{algorithmic}  
\end{algorithm}

\begin{equation}
\label{eq:modified_cost_function}
\min_{\Delta u_{k+j}} 
\sum_{i=1}^{N} \Vert x_{k+i} - x^{\rm ref}_{k+i} \Vert_{{\textit{\textbf{Q}}}}^2 + \sum_{j=0}^{N-1} \Vert \Delta{u_{k+j}} \Vert_{\textit{\textbf{R}}}^2
\end{equation}

\begin{equation}
\label{eq:modified_constrains}
\begin{aligned}
&\text{s.t.} \quad 
\begin{cases}
x_{k+i+1} = \Phi(x_{k+i}, u_{k+i},T,\hat \omega) \\
u_{k+i+1} = u_{k+i} + \sum_{j=0}^{i} \Delta u_{k+j} \\
u_{\min} \leq u_{k+i} \leq u_{\max} \\
\Delta u_{\min} \leq \Delta u_{k+i} \leq \Delta u_{\max} \\
x_{\min} \leq x_{k+i} \leq x_{\max}\\
\end{cases}, \\
&\quad\quad\quad \forall i = 0,\ldots,N-1.
\end{aligned}
\end{equation}

The NMPC above is numerically solved through the OSQP solver \cite{8516834}, and then the optimal control solution $u_k^{\rm opt}$ of NMPC is combined with PID control $u_k^{\rm pid}$ that synthesizes the composite controller (as in \cite{jin2022unknown}) addressing the tracking control of quadruped locomotion. 

\vspace{-0.1cm}

\section{Experiment}
In this section, hardware experiments are conducted to verify the efficacy of the proposed method on our robot platform.

\begin{figure}[ht]
    \centering
    \setlength{\abovecaptionskip}{-0.4 cm}
    \setlength{\belowcaptionskip}{-5 cm}
    \includegraphics[width=1\linewidth]{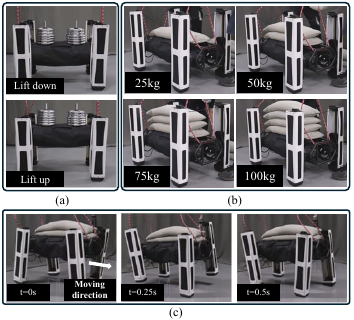}
    \caption{(a) shows the initial and terminal status of Kirin for payload identification. (b) represents the experiment scenario showing payloads of varying masses from 25 kg to 100 kg deployed on Kirin. (c) demonstrates the trotting test with 50kg payload.}
    \label{fig:robot_collection_new}
\end{figure}

\vspace{-0.4cm}
\subsection{Experiment Setup}
To validate and evaluate the performance of OPI-PIINC, tests are conducted on a real quadruped robot named Kirin which is introduced in our previous work \cite{luo2022prismatic}. The quadruped robot is electrically actuated with prismatic knee joints, which enable the robot to bear heavier payloads. Kirin has 12 degrees-of-freedom and weighs around 50 kg without payload and can be loaded with up to maximum of 225 kg payload. The prismatic joint is driven by quasi-direct drive (QDD) with 20:1 gear ratio. Such design makes the leg light enough compared with the total weight of the robot. Prismatic QDD is a trade-off solution to balance between the proprioceptive joints, impact mitigation, and high-bandwidth physical interaction.
\vspace{-0.3cm}
\subsection{PINNs Training}
\label{sec:PINN_Training}

The proposed PINNs architecture consists of a 4-layer multilayer perceptron (MLP) with 96 neurons per hidden layer, activated by ReLU functions. The network is trained using a hybrid optimization strategy, similar to the method in \cite{ramp-net}. Initial convergence is accelerated via the Adam optimizer with a learning rate of $1 \times 10^{-3}$ for 5000 epochs, followed by fine-tuning using the L-BFGS quasi-Newton method for enhanced precision. 

A Gazebo simulation platform is adopted to generate the training data for the quadruped robot under varying payload conditions, as shown in Fig.~\ref{fig:block_diagram}. Payloads ranging from 25 kg to 100 kg are applied to the robot’s torso, with a uniform sampling throughout the range to ensure comprehensive coverage of payload conditions. Sensor data containing joint angles, joint torques, robot linear/angular position, and ground reaction forces of each legs, are recorded at a sampling rate of 100 Hz during trotting tasks.  

A total of 100 distinct payload instances are generated, with each instance comprising 10-second motion sequences. Following the methodology outlined in \cite{antonelo2024physics}, the training dataset is composed of a series of input-output pairs $(\{x_0^i, u^i, T \} ,x^i)$, where the inputs are augmented to include initial states $x_0^i$, control inputs $u^i$, and time variable $T$ obtained by the above simulation. The selection of adjustable parameters is given as: ${\textit{\textbf{W}}=0.6{\textit{\textbf{I}}}_3}$, ${\textit{\textbf{V}}=0.8{\textit{\textbf{I}}}_3}$, ${\textit{\textbf{K}}=1.7{\textit{\textbf{I}}}_3}$. The training performance of PINNs exhibits that the total loss function eventually converges to $1 \times 10^{-4}$.
\vspace{-0.2cm}
\subsection{OPI-PINNPC for Quadruped Locomotion}

\begin{figure}[!htb]
    \centering
    \setlength{\abovecaptionskip}{-0.2 cm}
    \setlength{\belowcaptionskip}{-20 cm}
    \hspace{-2mm}
    \subfigure[]{
    \label{fig:pos_error}
    \includegraphics[width=0.492\linewidth]{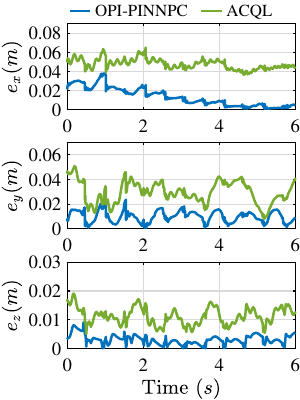}}
    \hspace{-3mm}
    \subfigure[]{
    \label{fig:ori_error}
    \includegraphics[width=0.492\linewidth]{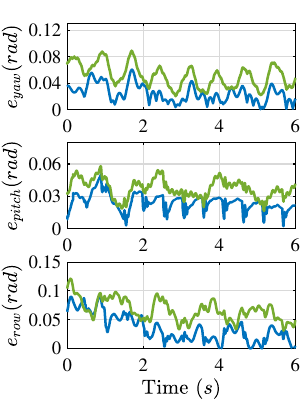}}
    \caption{Position and orientation tracking error curves of trotting test under 50kg payload with OPI-PINNPC and ACQL. (a) demonstrates the position tracking errors along the x-, y-, and z-axes, respectively. (b) shows the orientation tracking errors in yaw, pitch, and roll angles, respectively.}
    \label{fig:pos_ori_error}
\end{figure}
\begin{figure}[!htb]
    \centering
    \setlength{\abovecaptionskip}{-0.2 cm}
    \setlength{\belowcaptionskip}{-20 cm}
    \hspace{-2mm}
    \subfigure[]{
    \label{fig:pos_error_comp}
    \includegraphics[width=0.492\linewidth]{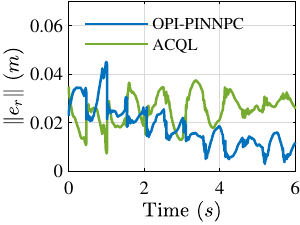}}
    \hspace{-3mm}
    \subfigure[]{
    \label{fig:ori_error_comp}
    \includegraphics[width=0.492\linewidth]{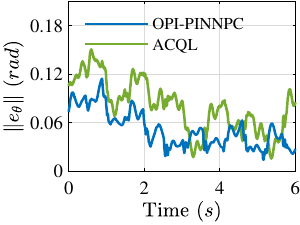}}
    \caption{Position and orientation tracking error norm curves of trotting test under 50kg payload with OPI-PINNPC and ACQL. (a) presents the comparison position tracking error norm curves. (b) shows the comparison orientation tracking error norm curves.}
\end{figure}
\begin{figure}[!htb]
    \centering
    \setlength{\abovecaptionskip}{-0.2 cm}
    \setlength{\belowcaptionskip}{-20 cm}
    \hspace{-5mm}
    \subfigure[]{
    \label{fig:pos_rmse_comp}
    \includegraphics[width=0.415\linewidth, angle=90]{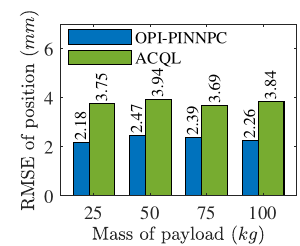}}
    \hspace{-3mm}
    \subfigure[]{
    \label{fig:ori_rmse_comp}
    \includegraphics[width=0.51\linewidth]{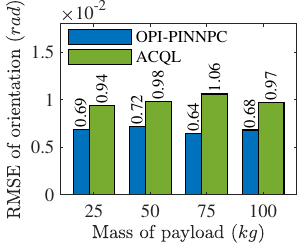}}
    \caption{RMSE of position and orientation with OPI-PINNPC and ACQL across payload mass variations (25-100 kg). (a) demonstrates the comparison RMSE of position. (b) presents the comparison RMSE of orientation.}
    \label{fig:rmse}
\end{figure}

In this section, a trotting test is performed on the quadruped robot to evaluate the effectiveness of OPI-PINNPC, with ACQL proposed in \cite{jin2022unknown} as the baseline. To validate the quadruped locomotion control performance under unknown payload conditions, the robot is configured to execute forward trotting test at a constant speed with a 50 kg payload. The experiment scenarios are shown in Fig.~\ref{fig:robot_collection_new} (d) with a time interval of 0.25 $s$. Specifically, the position tracking reference target is set to $(0.2 \cdot t(m), 0(m), 0.38(m))$ along the x-, y-, and z-axes, respectively, while the orientation tracking reference is defined as $(-0.05 (rad), (2.95 (rad), 0 (rad))$ for the pitch, yaw, and roll angles, respectively.

To highlight the advantages of the proposed OPI-PINNPC algorithm, we benchmark the performance of the baseline ACQL. Fig.~\ref{fig:pos_error} includes three subfigures corresponding to the position tracking errors along the x-, y-, and z-axes during the trotting test, while Fig. \ref{fig:ori_error} presents three subfigures depicting the orientation tracking errors in yaw, pitch, and roll angles respectively. The experimental results demonstrate that, compared with ACQL, the proposed OPI-PINNPC is able to guarantee higher accuracy in each dimension of position and orientation errors. 

For a clear differentiation between the two algorithms, the position and orientation error norm curves of the quadruped robot are employed to visualize the control performance. Specifically, Fig.~\ref{fig:pos_error_comp} and Fig.~\ref{fig:ori_error_comp} illustrate the position and orientation tracking error norm curves during the trotting test, respectively. The comparative results demonstrate that OPI-PINNPC achieves superior performance relative to ACQL, exhibiting higher tracking accuracy and a faster convergence rate throughout the experiments. 

Meanwhile, as evidenced by the experimental results in Fig.~\ref{fig:rmse}, the proposed OPI-PINNPC demonstrates adaptive control accuracy across the payload masses from 25 to 100 kg, achieving on average about 35\% higher precision than ACQL in position and orientation tracking performance.

\section{Conclusion and Future work}
\vspace{0.2cm}
The proposed OPI-PINNPC framework establishes a novel paradigm for locomotion control of quadruped robots under unknown payload conditions by integrating online payload identification with physics-informed learning. Due to the embedding of the OPI module that formulates the physics-informed loss functions through identified payload parameters in the training process, the composite loss function is ensured to rapidly converge, enabling accurate prediction for quadruped robot dynamics. Furthermore, a series of comparison experiments between OPI-PINNPC and ACQL reveal a 35\% improvement in trajectory tracking precision under payload conditions spanning 25–100 kg. This performance improvement can be attributed to the physics-informed loss function that explicitly embeds identified payload parameters into the PINN training process, effectively bridging data-driven learning with physical principles. Our future work will include the verification on versatile robot platform, such as robot manipulators.

\bibliographystyle{IEEEtran}
\bibliography{8-mybib}

\end{document}